\documentclass{article}
\usepackage{spconf,amsmath,graphicx,hyperref,booktabs,multirow}
\usepackage{algpseudocode}
\usepackage[ruled,vlined,linesnumbered]{algorithm2e}
\usepackage{array}

\usepackage{amssymb}
\usepackage{graphicx}   
\usepackage{subcaption} 
\usepackage{caption}   
\usepackage{float}  
\usepackage[dvipsnames]{xcolor}
\usepackage{colortbl}

\title{Fast Low-light Enhancement and Deblurring for 3D Dark Scenes}
\name{
Feng Zhang$^1$, Jinglong Wang$^1$, Ze Li$^2$, Yanghong Zhou$^3$, Yang Chen$^3$, Lei Chen$^1$\thanks{Lei Chen is the corresponding author}, Xiatian Zhu$^4$
}
\address{
$^{1}$ Nanjing University of Posts and Telecommunications, Nanjing. \\
$^{2}$ The Hong Kong University of Science and Technology, Hong Kong. \\
$^{3}$ The Hong Kong Polytechnic University, Hong Kong.
$^{4}$ University of Surrey, United Kingdom. 
}

\begin{document}
%
\maketitle
\begin{abstract}

Novel view synthesis from low-light, noisy, and motion-blurred imagery remains a valuable and challenging task.
Current volumetric rendering methods struggle with compound degradation, and sequential 2D preprocessing introduces artifacts due to interdependencies. 
In this work, we introduce FLED-GS, a fast low-light enhancement and deblurring framework that reformulates 3D scene restoration as an alternating cycle of enhancement and reconstruction.
Specifically, FLED-GS inserts several intermediate brightness anchors to enable progressive recovery, preventing noise blow-up from harming deblurring or geometry. 
Each iteration sharpens inputs with an off-the-shelf 2D deblurrer and then performs noise-aware 3DGS reconstruction that estimates and suppresses noise while producing clean priors for the next level. 
Experiments show FLED-GS outperforms state-of-the-art LuSh-NeRF, achieving 21× faster training and 11× faster rendering.

\end{abstract}
\section{Introduction}
\label{sec:intro}

Synthesizing novel views from low-light imagery degraded by sensor noise and motion blur is a challenging task with clear practical value for nighttime autonomous driving, robotic navigation in dim environments, and immersive VR.
Despite major advances in Neural Radiance Fields (NeRF)~\cite{mildenhall2021nerf} and 3D Gaussian Splatting (3DGS)~\cite{kerbl20233d} for novel view synthesis, these models struggle to recover well-lit, sharp, and clean views when low visibility, high-ISO noise, and camera shake co-occur in the inputs.

While applying sequential 2D enhancement methods before 3D reconstruction is straightforward, this simple combination is insufficient due to the complex interdependencies that cause cumulative artifacts. 
Several 3D reconstruction methods have been proposed to address either low-light scenes~\cite{li2025robust, cui2024aleth, zhou2025lita, cui2025luminance} or motion blur~\cite{pumarola2021d, lee2023dp, wang2023bad, lee2024deblurring} in isolation, yet they fail to account for the compound effects of these degradation factors. 
Recently, LuSh-NeRF~\cite{qu2024lush} introduced a collaborative framework for joint degradation removal during 3D reconstruction. 
However, this joint optimization suffers from slow convergence and becomes unreliable with similar blur patterns across views. Moreover, its multi-ray sampling per pixel to estimate camera trajectory imposes substantial computational overhead, preventing real-time deployment.

In this paper, we present FLED-GS, a fast low-light enhancement and deblurring framework for novel view synthesis from dark multi-view inputs. 
Direct enhancement from low-light conditions to target brightness not only amplifies noise excessively but also destroys high-frequency detail information essential for deblurring and geometric reconstruction. 
Based on this observation, we reformulate low-light scene restoration as an alternating cycle of enhancement and reconstruction.
Specifically, we establish several intermediate brightness levels as "anchors" between the low-light observations and desired illumination range, enabling progressive recovery that prevents excessive noise amplification from interfering with deblurring and reconstruction processes.
In each iteration, we utilize an existing 2D deblurring network to rapidly sharpen images, explicitly decoupling deblurring from denoising.
Subsequently, we perform noise-aware scene reconstruction based on 3DGS, explicitly estimating and suppressing noise in current brightness level. The reconstructed clean views serve as deblurring priors for the next brightness level. Through task decoupling and progressive recovery, FLED-GS ensures novel view quality while significantly reducing training and convergence costs.

\begin{figure*}[t]
    \centering
    \includegraphics[scale=0.33]{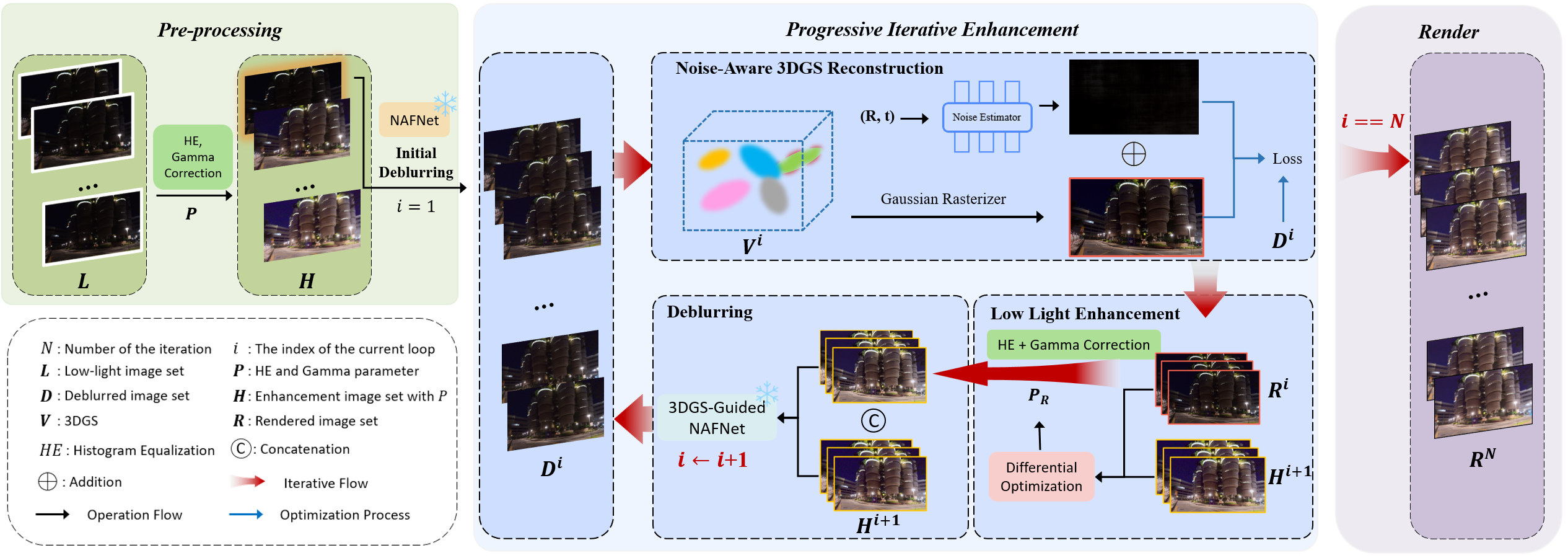} 
    \caption{ Overview of the pipeline for \textbf{FLED-GS}. Our framework consists of three main stages: Pre-processing, Progressive Iterative Enhancement, and Rendering. In the pre-processing stage, low-light images are enhanced step-by-step with varying histogram equalization and Gamma parameters to achieve progressive brightness adjustment. In the progressive iterative enhancement stage, the framework progressively performs low-light enhancement, deblurring, and denoising. Finally, the framework finishes training and leverages sharp normal-light 3DGS to render novel-view images. } 
    \label{fig:pipeline}
\end{figure*}

We summarize the contributions as follows:

(1) We introduce FLED-GS, to our knowledge the first 3DGS-based framework for novel view synthesis in low-light scenes with camera motion blur.

(2) We reformulate 3D restoration as an iterative cycle that alternates enhancement and reconstruction. The design decouples tasks by performing fast low-light enhancement and 2D deblurring, followed by noise-aware scene reconstruction with a lightweight noise estimator for dual-stage suppression.

(3) Extensive experiments demonstrate the superior performance of our method. Compared to the state-of-the-art LuSh-NeRF, FLED-GS achieves 21× faster training and 11× faster rendering.

\section{Methods}
\label{sec:methods}

\subsection{Progressive Iterative Enhancement Framework}
\label{sec:pie}

From captured low-light images, we observe two key phenomena: (1) low-light enhancement amplifies noise proportionally to enhancement intensity; (2) noise obscures camera shake trajectories, while preemptive denoising erases high-frequency motion cues, and heavy noise impedes deblurring. 
These insights lead to our "enhance-deblur-denoise" pipeline embedded within a progressive iterative 3D enhancement framework.

As illustrated in Figure~\ref{fig:pipeline}, our approach interpolates $N$ intermediate illumination levels between the observed and target brightness, executing $N+1$ optimization cycles. Each cycle comprises three key steps: low-light enhancement, deblurring, and noise-aware scene reconstruction. This progressive strategy effectively suppresses excessive noise amplification, laying a robust foundation for subsequent deblurring and reconstruction tasks.

The progressive iterative enhancement algorithm is presented in Algorithm~\ref{alg:pie}. At iteration $i$, parameter set $P_i$ enhances the low-light image $L$ to generate $H^i$. When $i=1$, NAFNet~\cite{chen2022simple} deblurs $H^1$ to produce $D^1$; for $i \geq 2$, $HR^{i}$ and $H^{i}$ are concatenated and processed by RF-guided NAFNet~\cite{choi2025exploiting} to generate $D^{i}$. The deblurred images $D^i$ are processed with noise-aware 3DGS to construct $V^i$, which renders images $R^i$. Differential optimization is performed to match $R^i$ with $H^{i+1}$ by searching for optimal parameters $\alpha \in [0, 15]$ and $\gamma \in [0.3, 1]$, yielding $P_R$. We employ Differential Evolution with population size 10, maximum 30 iterations, and early stopping (tolerance $= 10^{-4}$), minimizing a combined loss of MSE, SSIM, and histogram correlation. Using the optimized $P_R$, $R^i$ is enhanced to produce $HR^{i+1}$, which guides subsequent deblurring. This process iterates until $i = N$, resulting in a sharp radiance field $V^N$ for high-quality novel-view rendering.

\subsection{Noise-Aware 3DGS}
\label{ssec:noise}

Although the progressive iterative enhancement framework significantly suppresses noise, residual noise in rendering results remains inevitable. We propose a noise-aware module to enhance denoising capability. 
This module estimates the noise field using a 4-layer MLP with 128 hidden units per layer and an additional noise color output layer. The module samples spatial points based on the viewpoint camera pose $(R, T)$ as MLP inputs to construct a spatial noise field and generate the noise map for rendering.

 The noise map is calculated as:
\vspace{-0.1cm}
\begin{equation}
\small
I_{\mathrm{Noise}} = \mathrm{MLP}_{\mathrm{noise}}(\mathbf{x}, \mathbf{d})
\end{equation}

\noindent where $\mathbf{x}$ represents the spatial position of sampled points, and $\mathbf{d}$ denotes the viewing direction.

To guide scene reconstruction and noise estimation, we employ the following reconstruction loss, optimized via backpropagation:

\vspace{-0.1cm}
\begin{align}
\small
\mathcal{L}_{\mathrm{loss}} = (1 - \lambda)  \mathcal{L}_1&(I_{\mathrm{pred}},  I_{\mathrm{GT}}) 
+ \lambda \mathcal{L}_{\mathrm{ssim}}(I_{\mathrm{pred}}, I_{\mathrm{GT}}), \\
& I_{\mathrm{pred}} = I_{\mathrm{Noise}} \oplus I_r
\end{align}

\noindent where \(I_{\mathrm{GT}}\) represents the ground truth image, \(\lambda\) is set to 0.2, \(\oplus\) denotes the element-wise addition between the noise map \(I_{\mathrm{Noise}}\) and the 3DGS rendered image \(I_r\), and \(I_{\mathrm{pred}}\) is the predicted image.

\begin{algorithm}
\caption{Progressive Iterative Enhancement Framework}
\label{alg:pie}
\footnotesize
\KwIn{Low-light image \(L\); \newline Initial parameters \(P_0 = \{\alpha_0 = 0.1, \gamma_0 = 1\}\); \newline Final parameters \(P_N = \{\alpha_N, \gamma_N\}\); \newline Number of iterations \(N\)}
\KwOut{Rendered image \(R^N\)}
\BlankLine

\textbf{Step 1: Enhancement for All Iterations} \\
\(P \gets \textbf{logInterpolation}(P_0, P_N, N)\) \\
\(\{H^1, ..., H^N\} \gets \textbf{lowLightEnhanced}(L, P)\) 

\textbf{Step 2: Initial Deblurring} \\
\(D^1 \gets \textbf{initialDeblur}(H^1)\) 

\For{\(i = 1\) \textbf{to} \(N\)}{
  \textbf{Step 3: Noise-Aware Radiance Field Construction} \\
  \(V^i \gets \textbf{3dgsReconstruction}(D^i)\) \\
  \(R^i \gets \textbf{render3DGS}(V^i)\) 

  \textbf{Step 4: Low-Light Enhancement} \\
  \(P_R \gets \textbf{searchParams}(R^i, H^{i+1}, \alpha \in [0, 15], \gamma \in [0.3, 1])\) \\
  \(HR^{i+1} \gets \textbf{lowLightEnhanced}(R^i, P_R)\) 

  \If{\(i \neq N\)}{
    \textbf{Step 5: Deblurring} \\
    \(D^{i+1} \gets \textbf{3dgsGuidedDeblur}(HR^{i+1}, H^{i+1})\) 
  }
}
\textbf{Step 6: Final Rendering} \\
\(R^N \gets \textbf{render3DGS}(V^N)\) 

\Return \(R^N\)
\end{algorithm}

\begin{table*}[h]
    \centering
    \scriptsize
    \setlength{\tabcolsep}{0.5pt}
    \caption{Quantitative comparison of different methods on synthetic scenes of the  LuSh-NeRF dataset~\cite{qu2024lush}. \colorbox[HTML]{FFBDBD}{\makebox[1cm][c]{\phantom{A}}}: The best result; \colorbox[HTML]{FFF7B9}{\makebox[1cm][c]{\phantom{A}}}: The second best result.}
    \label{tab:result}
    \begin{tabular}{c|c c c|c c c|c c c|c c c|c c c} \toprule
        \multirow{2}{*}{Methods}      & \multicolumn{3}{c}{\textbf{“plane”}}                & \multicolumn{3}{c}{\textbf{“poster”}}               & \multicolumn{3}{c}{\textbf{“sakura”}}                 & \multicolumn{3}{c}{\textbf{“hall”}}                  & \multicolumn{3}{c}{\textbf{average}}  \\ \cmidrule{2-16}
                                & PSNR$\uparrow$ & SSIM$\uparrow$ &LPIPS$\downarrow$       & PSNR$\uparrow$ & SSIM$\uparrow$ &LPIPS$\downarrow$     & PSNR$\uparrow$ & SSIM$\uparrow$ &LPIPS$\downarrow$    & PSNR$\uparrow$ & SSIM$\uparrow$ &LPIPS$\downarrow$     & PSNR$\uparrow$ & SSIM$\uparrow$ &LPIPS$\downarrow$ \\ \midrule
3DGS~\cite{kerbl20233d}          & 5.57 & 0.0502 & 0.8899    & 11.33 & 0.1750 & 0.3693 & 7.64 & 0.1125 & 0.7757  & 8.18 & 0.1127 & 0.6379  & 8.18 & 0.0875 & 0.6682   \\ \midrule
\multicolumn{16}{c}{Image Enhancement Methods + Image Deblurring Methods + 3DGS} \\ \midrule

SCI~\cite{ma2022toward}+MLWNet~\cite{gao2024efficient}+3DGS~\cite{kerbl20233d}               
                                & 9.06 & 0.0489 & 0.5389 & 21.04 & 0.7266 & 0.1458 & 10.45 & 0.1915 & 0.4569 & 13.97 & 0.2222 & 0.3277 & 13.63 & 0.2729 & 0.3673 \\
                                
Zero-Dce~\cite{guo2020zero}+MLWNet~\cite{gao2024efficient}+3DGS~\cite{kerbl20233d}          
                                & 10.31 & 0.0197 & 0.5011 & 17.22 & 0.5670 & 0.1879 & 11.69 & 0.2266 & 0.4002 & 16.31 & 0.2754 & 0.3045 & 13.88 & 0.2623 & 0.3484 \\
Retinexformer~\cite{cai2023retinexformer}+Restormer~\cite{zamir2022restormer}+3DGS~\cite{kerbl20233d}  
                                & 16.73 & 0.3171 & 0.4662 & 17.43 & 0.5606 & 0.2290 & 17.51 & 0.4419 & 0.3487 & 23.11 & 0.6049 & 0.3226 & 18.69 & 0.4811 & 0.3416 \\
\midrule
\multicolumn{16}{c}{Image Enhancement Methods + Deblurring 3DGS} \\ \midrule
Zero-Dce~\cite{guo2020zero}+Deblurring-3DGS~\cite{lee2024deblurring}
                                & 9.81 & 0.0297 & 0.6653 & 7.97 & 0.1186  & 0.4580  & 9.16 & 0.1498 & 0.6683 & 14.04 & 0.1863 & 0.3459 & 10.24 & 0.1211  & 0.5344  \\
SCI~\cite{ma2022toward}+Debluring-3DGS~\cite{lee2024deblurring}                
                                & 7.83 & 0.0346 & 0.7274 & 21.75 & 0.7515 & 0.1467 & 8.19 & 0.1360 & 0.7933 & 12.97 & 0.1821 & 0.3696 & 12.68 & 0.2761 & 0.5093 \\
RetinexFormer~\cite{cai2023retinexformer}+DDRF~\cite{choi2025exploiting}              
                                & 17.19 & 0.3786 & 0.3176 & 18.25 & 0.6092 & 0.1225 & 16.68 & 0.4133 & 0.2974 & 21.98 & 0.6402 & \colorbox[HTML]{FFF7B9}{0.2624} & 18.52 & 0.5103 & 0.2500 \\
\midrule
\multicolumn{16}{c}{Image Deblurring Methods + Low Light Enhancement 3DGS} \\ \midrule
Restormer~\cite{zamir2022restormer}+LITA-GS~\cite{zhou2025lita}             
                                & 17.71 & 0.4134 & 0.4430 & 21.64 & 0.7587 & 0.1894 & 19.72 & 0.5757 & 0.3580 & 10.64 & 0.4379 & 0.6901 & 17.43 & 0.5464 & 0.4201 \\
MLWNet~\cite{gao2024efficient}+LITA-GS~\cite{zhou2025lita}                
                                & 16.89 & 0.4050 & 0.4369 & 21.43 & 0.7800 & 0.1787 & 18.58 & 0.5145 & 0.4496 & 17.71 & 0.5324 & 0.3873 & 18.65 & 0.5580 & 0.3631 \\
\midrule
\multicolumn{16}{c}{End-to-end Methods} \\ \midrule
LuSh-NeRF~\cite{qu2024lush} & 19.34 & 0.5375 & 0.3852 & 18.12 & 0.6331 & 0.2265 & 18.94 & 0.5884 & 0.2562 & 21.09 & \colorbox[HTML]{FFBDBD}{0.6421} & \colorbox[HTML]{FFBDBD}{0.2400} & 19.37 & 0.6003 & 0.2770 \\ \midrule
\textbf{FLED-GS(1 rounds) } & 20.63 & 0.4877 & 0.4222 & 24.30 & 0.7990 & 0.1259 & 19.84 & 0.5782 & 0.2572 & 23.93 & \colorbox[HTML]{FFF7B9}{0.6414} & 0.3136 & 22.18 & 0.6266 & 0.2797 \\
\textbf{FLED-GS(2 rounds) }         & \colorbox[HTML]{FFBDBD}{21.47} & 0.5988 & 0.3008                  & \colorbox[HTML]{FFBDBD}{24.31} & \colorbox[HTML]{FFBDBD}{0.8178} & \colorbox[HTML]{FFBDBD}{0.1049} & \colorbox[HTML]{FFBDBD}{20.60} & \colorbox[HTML]{FFBDBD}{0.6525} & \colorbox[HTML]{FFF7B9}{0.2087} & \colorbox[HTML]{FFBDBD}{24.00} & 0.5611 & 0.3102 & \colorbox[HTML]{FFBDBD}{22.60} & \colorbox[HTML]{FFBDBD}{0.6576} & \colorbox[HTML]{FFF7B9}{0.2312} \\
\textbf{FLED-GS(3 rounds)}          & \colorbox[HTML]{FFF7B9}{21.32} & \colorbox[HTML]{FFBDBD}{0.6064} & \colorbox[HTML]{FFF7B9}{0.2987}   & \colorbox[HTML]{FFF7B9}{24.13} & \colorbox[HTML]{FFF7B9}{0.8101} & 0.1080 & \colorbox[HTML]{FFF7B9}{20.35} &\colorbox[HTML]{FFF7B9}{0.6355} & 0.2135 & \colorbox[HTML]{FFF7B9}{22.44} & 0.5190 & 0.3050 & 22.06 & 0.6428 & 0.2313 \\
\textbf{FLED-GS(4 rounds)}          & 21.16 &\colorbox[HTML]{FFF7B9}{0.6022} &\colorbox[HTML]{FFBDBD}{0.2866}   & 24.00 &0.8088 &\colorbox[HTML]{FFF7B9}{0.1075} & 20.31 & 0.6205 & \colorbox[HTML]{FFBDBD}{0.2033} & 23.59 & 0.5732 & 0.2835 & 
\colorbox[HTML]{FFF7B9}{22.27} & \colorbox[HTML]{FFF7B9}{0.6512} & \colorbox[HTML]{FFBDBD}{0.2202} \\

\midrule
        \end{tabular}
\end{table*}

\section{Experiments}
\label{sec:experiment}

\subsection{Dataset and Implementation Details}
\label{sec:dataset}

We evaluate on the LuSh-NeRF dataset~\cite{qu2024lush}, containing synthetic and real-world subsets (5 scenes each, 20-25 images per scene, $1120 \times 640$ resolution). The synthetic subset provides ground truth, while the real-world subset demonstrates practical applicability under low-light conditions with motion blur. We additionally evaluate on the extremely low-light LOM dataset~\cite{cui2024aleth} (5 scenes) and the ExBlur dataset~\cite{lee2023exblurf} (10 scenes with extreme motion blur) to assess robustness under severe degradation.

All experiments are performed on a single NVIDIA V100 GPU (16GB memory) with 10,000 iterations. The noise estimator employs Adam optimizer with initial learning rate $1 \times 10^{-4}$ and exponential decay ($\gamma = 0.95$). All baselines adopt their official implementations with default settings.

\vspace{-3mm}
\subsection{Experimental Results}
\label{sec:results}
To evaluate our method, we conducted four comparative experiments: (1) combining image-based low-light enhancement~\cite{ma2022toward, guo2020zero, cai2023retinexformer} and deblurring methods~\cite{gao2024efficient, zamir2022restormer} with 3DGS~\cite{kerbl20233d}; (2) pairing low-light image enhancement~\cite{ma2022toward, guo2020zero} with deblurring 3DGS~\cite{lee2024deblurring}; (3) integrating image-based deblurring methods~\cite{gao2024efficient, zamir2022restormer} with low-light enhancement 3DGS~\cite{gao2024efficient}; and (4) using the state-of-the-art 3DGS LuSh-NeRF~\cite{qu2024lush} for simultaneous low-light enhancement and deblurring. Performance was assessed using PSNR, SSIM, and LPIPS metrics for comprehensive evaluation.

\begin{figure}[hbtp]
    \centering
    \includegraphics[scale=0.60]{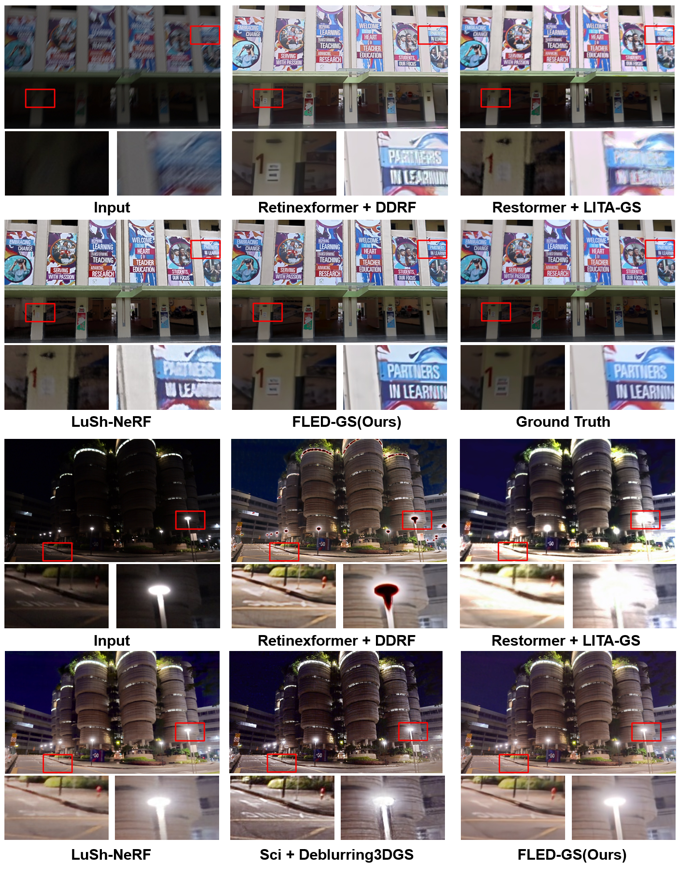} %
    \caption{Qualitative results of different methods on the LuSh-NeRF dataset. The upper scenes are synthetic scenes, and the lower scenes are real scenes.}
    \label{fig:qualitative} 
    \vspace{-4mm}
\end{figure}

Table~\ref{tab:result} demonstrates that FLED-GS consistently matches or exceeds the state-of-the-art methods in all metrics, with qualitative validation in Figure~\ref{fig:qualitative}. On the severely low-light LOM dataset~\cite{cui2024aleth}, we notably outperform LuSh-NeRF (Table~\ref{tab:lom-dataset}). However, our method struggles with extreme motion blur due to COLMAP failure. On the ExBlur dataset~\cite{lee2023exblurf}, COLMAP failed on 4 out of 10 scenes. On the remaining 6 scenes, we underperform the specialized ExBlurRF~\cite{lee2023exblurf}, suggesting COLMAP-free approaches as promising future work. Table~\ref{tab:ablation1} presents efficiency and complexity analysis. Our 3DGS-based model contains more parameters than LuSh-NeRF (12.17M vs 1.24M), but achieves 21$\times$ faster training (41min vs 14.5h) and 11$\times$ faster rendering (0.8s vs 9.1s), reflecting the classic 3DGS vs. NeRF trade-off.

\subsection{Ablation Study}
\label{sec:typestyle}
Table~\ref{tab:ablation2} presents ablation study results on synthetic scenes from the LuSh-NeRF dataset~\cite{qu2024lush}, with metrics averaged across all scenes. We select $N=2$ iterations for optimal PSNR/SSIM while maintaining training efficiency (see Table~\ref{tab:result}). The baseline (w/o PIE, w/o NE) achieves 21.65 PSNR, 0.5831 SSIM, and 0.2254 LPIPS. Incorporating PIE alone (w/ PIE, w/o NE) improves structural quality (21.94 PSNR, 0.6250 SSIM) but increases LPIPS to 0.2444, as PIE introduces residual noise to preserve motion cues for deblurring. Using only the Noise Estimator (w/o PIE, w/ NE) yields moderate improvements (21.79 PSNR, 0.5885 SSIM, 0.2372 LPIPS). The full FLED-GS configuration combining both components achieves the best performance (22.60 PSNR, 0.6576 SSIM, 0.2312 LPIPS), demonstrating that the Noise Estimator effectively recovers perceptual quality degraded by PIE's iterative process, validating our design rationale. 

\begin{table}[h]
    \scriptsize
    \centering
    \caption{Quantitative results on the extremely low-light LOM dataset~\cite{cui2024aleth}, averaged across all scenes.}
    \label{tab:lom-dataset}
    \setlength{\tabcolsep}{11pt}
    \begin{tabular}{cccc} 
        \toprule	
        \textbf{Method}  & PSNR$\uparrow$ & SSIM$\uparrow$ & LPIPS$\downarrow$ \\ \midrule 
        Retinexformer~\cite{cai2023retinexformer}+3DGS~\cite{kerbl20233d}     & 18.33  & 0.755  & 0.359  \\	
        LITA-GS~\cite{zhou2025lita}                & 18.04  & 0.728  & 0.369  \\
        Aleth-NeRF~\cite{cui2024aleth}             & 19.87  & 0.754  & 0.417  \\
        LuSh-NeRF~\cite{qu2024lush}                & 21.03  & 0.726  & 0.265  \\
        \textbf{FLED-GS(2 rounds)}                 & 22.77  & 0.824  & 0.205  \\
        \bottomrule 
    \end{tabular}
\end{table}

\begin{table}[h]
    \centering
    \scriptsize
    \caption{Efficiency and complexity comparison. Rendering time refers to the average time required to render a single image. Model parameters are averaged across scenes.}
    \label{tab:ablation1}
    \setlength{\tabcolsep}{5.5pt}
    \begin{tabular}{ccccc} 
        \toprule
        \textbf{Methods}  & Training time & Rendering time & Model Parameters \\ \midrule 
        LuSh-NeRF~\cite{qu2024lush}          & 14h30min & 9.1s & 1.24M  \\	\midrule
        \textbf{FLED-GS(2 rounds)}     & 41min & \multirow{3}{*}{0.8s} & \multirow{3}{*}{12.17M} \\
        \textbf{FLED-GS(3 rounds)}     & 1h8min &  &  \\
        \textbf{FLED-GS(4 rounds)}     & 1h34min &  & \\
        \bottomrule 
    \end{tabular}
\end{table}

\begin{table}[h]
    \scriptsize
    \centering
    \caption{Effect of PIE and NE on synthetic scenes. The metric represents the average across synthetic scenarios.}
    \label{tab:ablation2}
    \setlength{\tabcolsep}{15.5pt}
    \begin{tabular}{cccc} 
        \toprule	
        \textbf{Configuration}  & PSNR$\uparrow$ & SSIM$\uparrow$ & LPIPS$\downarrow$ \\ \midrule 
        w/o PIE, w/o NE & 21.65  & 0.5831  & 0.2254  \\	
        w/ PIE, w/o NE  & 21.94 & 0.6250  & 0.2444  \\
        w/o PIE, w/ NE  & 21.79 & 0.5885 & 0.2372   \\
        Full FLED-GS        & 22.60 & 0.6576 & 0.2312   \\
        \midrule 
    \end{tabular}
\end{table}

\section{Conclusion}
\label{sec:conclusion}
In this work, we present FLED-GS, a fast framework for novel view synthesis in low-light scenes with camera motion blur. We reformulate 3D restoration as a lighting-anchored iterative cycle that alternates enhancement and reconstruction, decoupling tasks and curbing noise amplification. A dedicated noise estimator further enhances visual quality, yielding precise reconstructions. Extensive experiments confirm our approach outperforms state-of-the-art methods in visual fidelity and robustness across varied low-light conditions. 

\section{ACKNOWLEDGMENT}
\label{sec:acknowledgement}
This work is supported by the National Natural Science Foundation of China (Grant No. 62202241), Jiangsu Province Natural Science Foundation for Young Scholars (Grant No. BK20210586), NUPTSF (Grant No. NY221018) and Double-Innovation Doctor Program under Grant JSSCBS20220657.

\vfill\pagebreak

\bibliographystyle{IEEEbib}
\bibliography{strings,refs}

\end{document}